\documentclass[journal,transmag]{IEEEtran}

\ifCLASSINFOpdf
\else
\fi
\usepackage{times}
\usepackage{epsfig}
\usepackage{graphicx}
\usepackage{amsmath}
\usepackage{amssymb}
\usepackage{chngcntr}
\usepackage[]{algorithm2e}

\usepackage[utf8]{inputenc}
\usepackage[english]{babel}
\usepackage{comment}
\usepackage{amsthm}
\usepackage[usenames, dvipsnames]{color}

\usepackage[switch]{lineno}
 
\definecolor{Pink}{rgb}{0.858, 0.188, 0.478}
\DeclareMathOperator*{\amin}{min}
\DeclareMathOperator*{\amax}{max}

\begin{document}

\title{\textcolor{black}{Real-time Deep Pose Estimation with Geodesic Loss for Image-to-Template Rigid Registration}}

\author{\IEEEauthorblockN{Seyed Sadegh Mohseni Salehi\IEEEauthorrefmark{1,2},~\IEEEmembership{Student Member,~IEEE},
Shadab Khan\IEEEauthorrefmark{1},~\IEEEmembership{Member,~IEEE,}, \\ Deniz Erdogmus\IEEEauthorrefmark{2},~\IEEEmembership{Senior Member,~IEEE}, and
Ali Gholipour\IEEEauthorrefmark{1},~\IEEEmembership{Senior Member,~IEEE}}
\IEEEauthorblockA{\IEEEauthorrefmark{1}Radiology Department, Boston Children's Hospital; and Harvard Medical School, Boston MA 02115}
\IEEEauthorblockA{\IEEEauthorrefmark{2}Electrical and Computer Engineering Department, Northeastern University, Boston, MA, 02115}

\thanks{Copyright (c) 2017 IEEE. Personal use of this material is permitted. However, permission to use this material for any other purposes must be obtained from the IEEE by sending a request to pubs-permissions@ieee.org. Manuscript received March 9, 2018.
Corresponding author: S.S.M.Salehi (email: ssalehi@ece.neu.edu). This study was supported in part by the McKnight Foundation through a Technological Innovations in Neuroscience Award, and in part by the National Institute of Biomedical Imaging and Bioengineering of the National Institutes of Health (NIH) grant R01 EB018988. Relevant code can be found at: https://github.com/SadeghMSalehi/DeepRegistration}}

\maketitle
\begin{abstract}
With an aim to increase the capture range and accelerate the performance of state-of-the-art inter-subject and subject-to-template 3D rigid registration, we propose deep learning-based methods that are trained to find the 3D position of arbitrarily oriented subjects or anatomy \textcolor{black}{in a canonical space} based on slices or volumes of medical images. For this, we propose regression convolutional neural networks (CNNs) that learn to predict the angle-axis representation of 3D rotations and translations using image features. \textcolor{black}{We use and compare mean square error and geodesic loss to train regression CNNs for 3D pose estimation used in two different scenarios: slice-to-volume registration and volume-to-volume registration.} As an exemplary application, we applied the proposed methods to register arbitrarily oriented reconstructed images of fetuses scanned \textit{in-utero} at a wide gestational age range to a standard atlas space. Our results show that in such registration applications that are amendable to learning, the proposed deep learning methods with geodesic loss minimization \textcolor{black}{achieved 3D pose estimation} with a wide capture range in real-time ($<100ms$). We also tested the generalization capability of the trained CNNs on an expanded age range and on images of newborn subjects with similar and different MR image contrasts. We trained our models on T2-weighted fetal brain MRI scans and used them to predict the 3D pose of newborn brains based on T1-weighted MRI scans. We showed that the trained models generalized well for the new domain when we performed image contrast transfer through a conditional generative adversarial network. This indicates that the domain of application of the trained deep regression CNNs can be further expanded to image modalities and contrasts other than those used in training. A combination of our proposed methods with accelerated optimization-based registration algorithms can dramatically enhance the performance of automatic imaging devices and image processing methods of the future.

\end{abstract}


\IEEEdisplaynontitleabstractindextext

\IEEEpeerreviewmaketitle

\section{Introduction}

\subsection{Background}

\IEEEPARstart{I}{mage} registration is one of the most fundamental tools in biomedical image processing, with applications that range from image-based navigation in imaging and image-guided interventions to longitudinal and group analyses~\cite{hill2001medical,pluim2003mutual,gholipour2007brain,shams2010survey,markelj2012review,sotiras2013deformable,ferrante2017slice}. 
Registration can be performed between images of the same modality or across modalities, and within a subject or across subjects, with diverse goals such as motion correction, pose estimation, spatial normalization, and atlas-based segmentation. 
Image registration is defined as an optimization problem to find a global transformation or a deformation model that maps a source (or moving) image to a reference (or fixed) image. The complexity of the transformation is defined by its degree-of-freedom (DOF) or the number of its parameters. The most widely used transformations in biomedical image registration range from rigid and affine to high-dimensional small or large deformations based on biophysical/biomechanical, elastic, or viscous fluid models~\cite{sotiras2013deformable}.

Given a transformation model and images, iterative numerical optimization methods are used to maximize intensity-based similarity metrics or minimize point cloud or local feature distances between images; however the cost functions associated with these metrics are often non-convex, limiting the capture range of these registration methods. Techniques such as center-of-gravity matching, principal axes and moments matching, grid search, and multi-scale registration are used to initialize transformation parameters so that iterative optimization starts from the vicinity of the global optimum. These techniques, however, are not always successful, especially if the range of possible rotations is wide and shapes have complex features. Grid search and multi-scale registration may find global optima but are computationally expensive and may not be useful in time-sensitive applications such as image-based navigation.

There has been an increased interest in using deep learning in medical image processing, motivated by promising results that have been achieved in semantic segmentation in computer vision~\cite{long2015fully} and medical imaging~\cite{greenspan2016guest,litjens2017survey}. The use of learning-based techniques in image registration, however, has been limited. Some registration tasks, for example those on image to template, atlas, or standard-space registration are amendable to learning and may provide significant improvement over strategies such as iterative optimization or grid search when the range of plausible position/orientation is wide, demanding a large capture range. Under these conditions, a human observer can find the approximate pose of 3D objects quickly and bring them into rough alignment without solving an iterative optimization. This is performed through feature identification.

\subsection{Related Work}

Deep feature representations have recently been used to learn metrics to guide local deformations for multi-modal inter-subject registration~\cite{simonovsky2016deep,wu2016scalable}. These works have shown that deep learned metrics provide slight improvements over local image intensity and patch features that are currently used in deformable image registration. Initialized by rigid and affine alignments, the goal here was merely to improve local deformations and not the global alignment. In another recent work on deformable registration, Yang et al.~\cite{yang2017quicksilver} developed a deep autoencoder-decoder convolutional neural network (CNN) that learned to predict the Large Deformation Diffeomorphic Metric Mapping (LDDMM) model, and achieved state-of-the-art performance with an order of magnitude faster optimization in inter-subject and subject-to-atlas deformable registration.

For 3D global rigid registration, which is the subject of this study, Liao et. al.~\cite{liao2017artificial} proposed a reinforcement learning algorithm for a CNN with 3 fully connected layers. They used a greedy supervised learning strategy with an attention-driven hierarchical method to simultaneously encode a matching metric and learn a strategy; and showed improved accuracy and robustness compared to state-of-the-art registration methods in computed tomography (CT). This algorithm is relatively slow and lacks a systematic stopping criterion at test time.

In an effort to speed up slice-to-volume (X-ray to CT) rigid registration and improve its capture range, Miao et. al.~\cite{miao2016real, miao2016cnn} proposed a real-time registration algorithm using CNN regressors. In this method, called pose estimation via hierarchical learning, they partitioned the 6-dimensions of the parameter space to three zones to learn, hierarchically, the regression function based on in-plane and out-of-plane rotations and out-of-plane translations. CNN regressors were trained separately in each zone, where local image residual features were used as input and the Euclidean distance of the transformation parameters were used as the loss function. In experiments with relatively small rotations of up to $30^{\circ}$ (perturbations with standard deviations of $10^{\circ}$ in each of the rotation parameters), they reported improved registrations achieved in 100ms ($\sim$20-45 times faster than the best intensity-based slice-to-volume registration in that application).

The slice-to-volume (X-ray to CT) image registration problem shares similarity with 3D pose estimation in computer vision. The term 3D pose estimation in computer vision is referred to as finding the underlying 3D transformation between an object and the camera from 2D images. State-of-the-art methods for CNN-based 3D pose estimation can be classified in two groups: 1) models that are trained and used to predict keypoints as models and then use object models to find the orientation~\cite{wu2016single, pavlakos20176}; and 2) models that predict the pose of the object directly from images~\cite{tulsiani2015viewpoints, su2015render}. Pose estimation in computer vision has been largely treated as a classification problem, where the pose space is discretized into bins and the pose is predicted to belong to one of the bins~\cite{tulsiani2015viewpoints,su2015render}. Mahendran et al.~\cite{mahendran20173d} have recently modeled the 3D camera/object pose estimation as a regression problem. 
They proposed deep CNN regression to find rotation matrices and a new loss function based on geodesic distance for training.


\subsection{Contributions}

Similar to~\cite{miao2016real,miao2016cnn,mahendran20173d}, we propose deep CNN regression models \textcolor{black}{for 3D pose estimation}; but unlike those works that focused on estimating pose based on 2D-projected image representation of objects (thus limited rotations), we aimed to find the 3D pose of arbitrarily-oriented objects based on their volumetric or sectional (slice) image representations. \textcolor{black}{In this paper, we use the term 3D pose mainly to refer to 3D orientation; and use registration for the estimation of both rotation and translation parameters in 3D.} Our goal was to speed up and improve the capture range of volume-to-volume and slice-to-volume registrations. \textcolor{black}{To achieve this, we formulated a regression problem for 3D pose estimation} based on the angle-axis representation of 3D rotations that form a Special Orthogonal Group $SO(3)$; and used the bi-invariant geodesic distance, which is a natural \textcolor{black}{Riemannian} metric on $SO(3)$~\cite{huynh2009metrics}, as the loss function. We augmented our proposed deep regression network with a correction network to estimate translation parameters, and ultimately used it to initialize optimization-based registration to achieve robust and accurate registration at the widest plausible range of 3D rotations. \textcolor{black}{In this paper we do not suggest a general method of registration for arbitrary pairs of images. Rather, 3D pose estimation finds the orientation of a shape or anatomy with respect to a canonical space or template. Inter-subject registration can then be achieved by computing a composite transformation from estimated 3D pose of images of individual subjects.}

We applied our proposed method to rigidly register reconstructed fetal brain MRI images~\cite{gholipour2010robust} to a standard (atlas) space. Fetal brains can be in any arbitrary orientation with respect to the MRI scanner coordinate system, as one cannot pre-define the position of a fetus when a pregnant woman is positioned on an MRI scanner table. Moreover, fetuses frequently move and can rotate within scan sessions. Our deep model, trained on reconstructed T2-weighted images of 28-37 week gestational age (GA) fetuses from the training set, was able to find the 3D position of fetuses in the test set in real-time ($<100$ms) in the majority of cases, where optimization-based methods failed due to falling in local minima. We then examined the generalization properties of the learned model on test images of much younger fetuses (21-27 weeks GA), as well as T2- and T1-weighted images of newborns, that all exhibited significantly different size, shape, and features.

Based on our formulation, we also trained models for slice-to-volume registration, an application that exhibits significant technical challenges in medical imaging, as recently reviewed in~\cite{ferrante2017slice}. Prior work on slice-to-volume registration in fetal MRI has shown a strong need for regularization and initialization of slice transformations through hierarchical registration~\cite{gholipour2010robust,kainz2015fast} or state-space motion modeling~\cite{marami2017temporal}. Learning-based methods have been recently used to improve prediction of slice locations in fetal MRI~\cite{hou2017predicting,hou20183d} and fetal ultrasound~\cite{namburete2018fully}. In~\cite{hou2017predicting,hou20183d} anchor-point slice parametrization was used along with the Euclidean loss function based on~\cite{kendall2015posenet} to predict slice positions and reconstruct fetal MRI in canonical space. The alignment of fetal ultrasound slices in~\cite{namburete2018fully} was formulated as z-position estimation and 3-class slice plane classification (mid-axial, target eye, and eye planes); where a CNN was trained using negative likelihood loss for simultaneous prediction of slice location and brain segmentation.

For slice-to-volume registration we used 3D full rotation representation to train our CNN regression model. Our results are also promising in this application as they show initial pose of the fetal head can be estimated in real time from slice acquisitions, which is particularly helpful if good-quality slices are only sparsely acquired due to fetal motion. \textcolor{black}{Real-time pose estimation and registration has broader potential applications such as guided and automated ultrasonography~\cite{namburete2018fully}, automated fetal MRI~\cite{hou2017predicting,gholipour2014fetal}, and motion-robust MRI~\cite{thesen2000prospective,white2010promo,gholipour2011motion,marami2016motion,kurugol2017motion}.} \textcolor{black}{For example Hou et al.~\cite{hou20183d} used slice-to-volume registration for fetal brain MRI reconstruction. Real-time slice-to-volume registration may also be used for real-time motion tracking in MRI of moving subjects for data re-acquisition or prospective navigation.} The remainder of this paper involves details of the methods in Section~\ref{sec:method}, followed by results in Section~\ref{sec:results}, a discussion in Section~\ref{sec:discussion}, and the conclusion. Our formulation is generic and may be used in other applications.


\section{Methods}
\label{sec:method}
In this section we present a 3D rotation representation that helps us build our CNN regression models for 3D pose estimation. We show how using a non-linear activation function can mimic exact rotation representation. We present our network architectures and propose a two-step training algorithm with appropriate loss functions to train the network.
\subsection{3D Rotation Representation}
\label{sec:Representation}
A 3D rotation is commonly represented by a $3 \times 3$ matrix $R$ with 9 elements that are subject to six norm and orthogonality constraints ($R$ is orthogonal and $\texttt{det}R=1$). The set of 3D rotations form the Special Orthogonal Group $SO(3)$ that is a 3-dimensional object embedded in $\mathbb{R}^4$ (thus has 3 DOFs). $SO(3)$ is a compact Lie group that has skew symmetric matrices as its Lie algebra. Its 3 DOFs can be represented as 3 consecutive rotations relative to principle axes of the coordinate frame.

Based on Euler's theorem each rotation matrix $R$ can be described by an axis of rotation and an angle around it (known as angle-axis representation). A 3-dimensional rotation vector is a compact representation of rotation matrix such that rotation axis is its unit vector and angle in radians is its magnitude. The axis is oriented so that the angle rotation is counterclockwise around it. As a consequence, the rotation angle is always non-negative, and at most $\pi$; i.e. $\theta\in[0,\pi)$.

For a 3-dimensional vector $v$, by defining $\hat{v} = \frac{v}{\left \| v \right \|_2}$ as the axis of orientation and $\theta = {\left \| v \right \|_2}$ as the angle of rotation (in radians), the rotation matrix is calculated as:
\begin{equation}
    R = exp(\theta[\hat{v}]_\times)
    \label{eq:exp}
\end{equation}
where $[{\hat{v}}]_\times$ is the skew-symmetric operator:
\begin{equation}
    [{\hat{v}}]_\times = \begin{bmatrix}
0 & -\hat{v}_3 & \hat{v}_2 \\ 
\hat{v}_3 & 0 &-\hat{v}_1 \\ 
 -\hat{v}_2& \hat{v}_1 & 0
\end{bmatrix}
\label{skew}
\end{equation}
\noindent Using Rodrigues' rotation formula, (\ref{eq:exp}) can be simplified to:
\begin{equation}
    R = I_3 + sin(\theta)[\hat{v}]_\times + (1-cos\theta)[\hat{v}]_\times^2
    \label{rod}
\end{equation}
\noindent and
\begin{equation}
    tr(R) = 1 + 2cos(\theta)
    \label{trace}
\end{equation}
\noindent As a result, to find any arbitrary rotation in 3D space it is sufficient to find the rotation vector $v$ corresponding to that orientation. In the next section, the proposed networks that can find this rotation vector are introduced. 

\begin{figure*}
    \centering
    \includegraphics[width=0.9\textwidth]{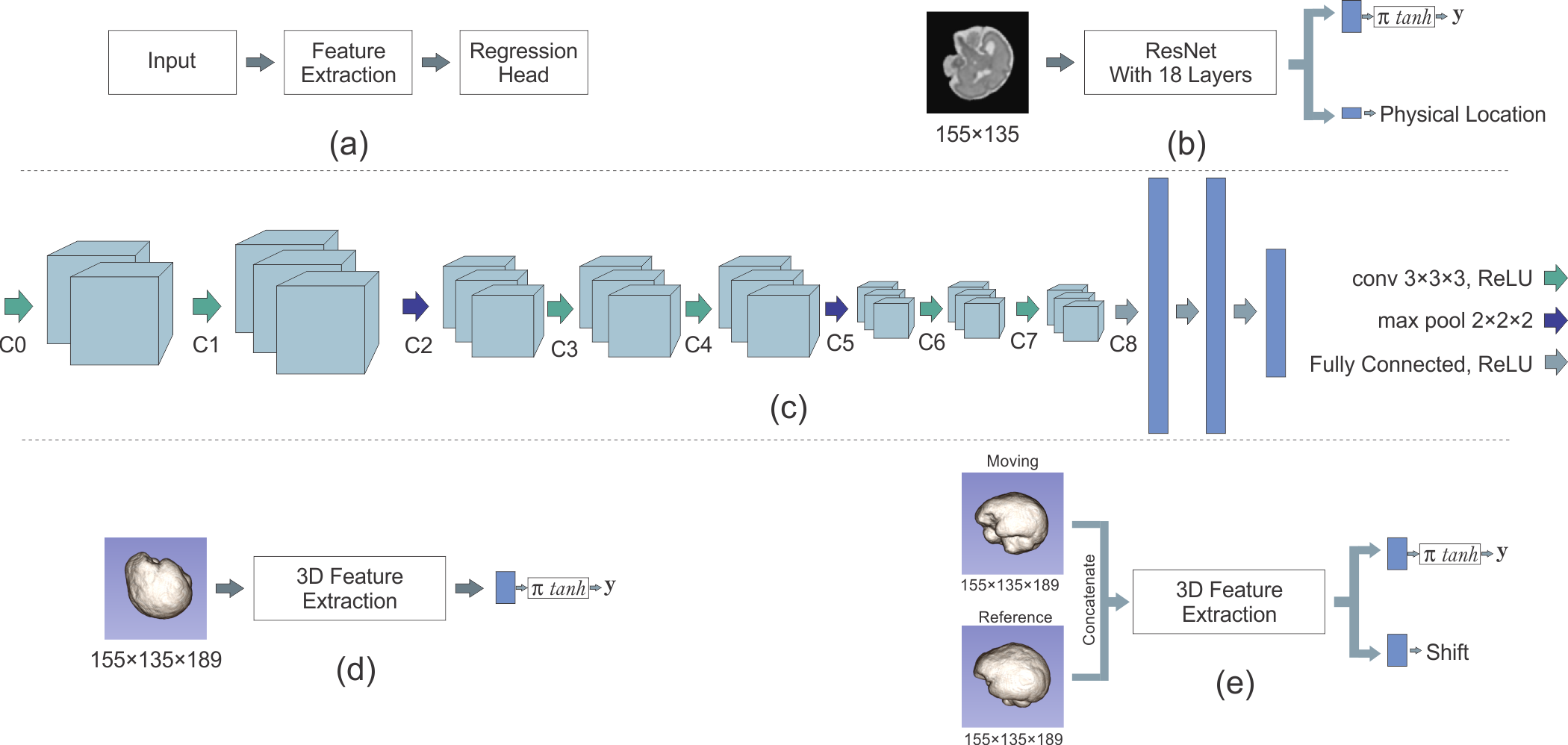}
    \caption{Schematic diagram of the proposed networks: (a) Different parts of a regression network; \textcolor{black}{(b) The proposed architecture of the slice 3D pose estimation network; (c) The feature extraction part of the volume 3D pose estimation network; (d) The proposed architecture of the volume 3D pose estimation network; and (e) The correction network used for the prediction of both rotation and translation parameters between moving and reference images.}}
    \label{fig:Net}
\end{figure*}

Figure~\ref{fig:Net}(a) shows general parts of the regression networks used in this study. Each network contains 3 parts: input, feature extraction, and output. In this study we used three networks with slightly different configurations of these parts. Next we discuss the architecture of each network in detail.   

\subsection{\textcolor{black}{Slice 3D Pose Estimation Network Architecture}}
\textcolor{black}{For slice 3D pose estimation we used an 18-layer residual CNN~\cite{he2016identity} for feature extraction, and \textcolor{black}{two regression heads\footnote{The top of the network is referred to as the head of the network.}}: one for regression over 3 rotation parameters and the other for slice location in the atlas space. }\textcolor{black}{While different choices existed for the feature extraction component of the networks, in choosing a network architecture for slice pose estimation we tried different network architectures based on suggestions from pose estimation literature, including~\cite{hou20183d}. We examined VGG16, ResNet18, and DenseNet. We observed better performance with ResNet18 compared to other networks including VGG16.} The network architecture is shown in Figure~\ref{fig:Net}(b). For the rotation head, the last fully connected layer has size three which corresponds to the elements of the rotation vector $v$. \textcolor{black}{The last non-linear function on top of the fully connected layer is $\pi \times tanh$ which limits the output of each element from $-\pi$ to $+\pi$ and simulates the constraints of each element of the rotation vector independently.} \textcolor{black}{The physical location estimator head contains a scalar, as this network tries to estimate the physical location of the slice (in millimeters) along with its orientation.} ReLU non-linearity is applied on top of this head as the value of the slice number is non-negative.

\subsection{\textcolor{black}{Volume 3D Pose Estimation Network Architecture}}
\textcolor{black}{
The 3D feature extraction part of our 3D pose estimation network for volume-to-volume registration is shown in Figure~\ref{fig:Net}(c), where block arrows show the functions defined on the right hand side of the figure.} All convolutional kernels have size $3 \times 3 \times 3$. In the first layer, eight convolutional kernels are applied on the 3D input image, followed by ReLU nonlinear function and batch normalization \textcolor{black}{(C0 and C1). The tensors are down-sampled by a factor \textcolor{black}{of} 2 using the 3D max-pooling function (C2) before the second and third convolutional layers (C3 and C4). For C3 and C4, ReLU nonlinear function and batch normalization are used after applying 32 convolutional kernels. In the last two convolutional layers (C6 and C7), 64 kernels are used followed by ReLU and batch normalization. Following C7, 3 fully connected layers with size of 512, 512, and 256 are used with ReLU nonlinear \textcolor{black}{activation} function and batch normalization (C8)}. The feature extraction part provides 256 features that are fed into the regression head. The overall architecture of the pose network is shown in Figure~\ref{fig:Net}(d). \textcolor{black}{This network estimates orientation, and has the same regression rotation head as the slice pose estimation network.}

\subsection{Volume-to-Volume Correction Network Architecture}
\textcolor{black}{The correction network aims at simultaneously estimating translations and rotations.} \textcolor{black}{Note that we assume initial translations between stack-of-slices or volumes and the template (or reference) are calculated by center-of-gravity matching and initial rotations are estimated by the volume 3D pose estimation networks, so the correction network aims to register a roughly-aligned source image to the reference template. The architecture of this network is shown in Figure~\ref{fig:Net}(e).} The 3D feature extraction part of this network is the same as the volume 3d pose estimation network. In this architecture, both a 3D reference image (an atlas or template image) and a roughly-oriented 3D moving image are fed as 2-channel input, as we aim to estimate both rotation and translation parameters. The regression head of this network contains two heads: a rotational head as already described and a translational head. The translational head is a vector of 3 parameters that translate the moving image into the target image.

\subsection{Training the Networks}
In this section we describe the training procedures for the networks. The loss function is designed as:
\begin{equation}
    Loss_{Total} = Loss_{Rotation} + \lambda Loss_{Translation}
    \label{eq:total}
\end{equation}
\noindent where $\lambda$ is a hyper-parameter to balance between the rotation loss $Loss_{Rotation}$ (which is bounded between 0 to $\pi$) and the translation loss $Loss_{Translation}$. The translation loss is the mean-squared error (MSE) between the predicted and ground truth translation vectors. For the first stage of training, we use the MSE loss also for the rotation parameters, and then switch to the geodesic loss in the second stage. The MSE loss is defined as
\begin{equation}
    Loss_{MSE} = \left \| v-y \right \|_2
    \label{eq:MSE}
\end{equation}
where $y$ and $v$ are the output of the rotation head and the ground truth rotation, respectively. 
\textcolor{black}{MSE, as a convex loss function, can help narrow down the search space for pose prediction learning, thus is appropriate for training; however, it does not accurately represent a distance function between two rotations.} The distance between two 3D rotations is geometrically interpreted as the geodesic distance between two points on the unit sphere. The geodesic distance (or the shortest path) is the radian angle between two viewpoints, which has an exponential form. 
\textcolor{black}{Let $R_{s}$ and $R_{GT}$ $\in SO(3)$ be the estimated and the ground truth rotation matrices, respectively. The distance between these rotation matrices is defined as:}
\begin{equation}
    d(R_s, R_{GT}):SO(3)\times SO(3) \rightarrow \mathbb{R}^{+} = \theta
    \label{eq:dis}
\end{equation}
Equation~(\ref{eq:dis}) shows the amount of rotation in radian around a specific vector that needs to be applied on rotation matrix $R_s$ to reach rotation matrix $R_{GT}$, and is calculated as:
\begin{equation}
    d(R_s, R_{GT}) = \left \| log(R_s^TR_{GT}) \right \|_F
    \label{eq:dis2}
\end{equation}
where $\left \| . \right \|_F $ is the Frobenius norm and $log(.)$ is the matrix logarithm of a rotation matrix that can be written as:
\begin{equation}
    log(R) = \left\{\begin{matrix}
0 & if \ \theta=0\\ 
\frac{\theta}{2sin(\theta)} (R-R^T)& if \ 0< \theta<\pi 
\end{matrix}\right. 
    \label{eq:logR}
\end{equation}
To show that (\ref{eq:dis2}) is actually the distance between rotation matrices we should consider the fact that a rotation matrix is orthogonal ($R^{-1}=R^T$) and the rotation from $R_s$ to $R_{GT}$ is $R_s^TR_{GT}$. Considering (\ref{eq:logR}) and the fact that $R-R^T = 2sin(\theta)[\hat{v}]_\times$ can be calculated using (\ref{rod}), where $\hat{v} = \frac{v}{\left \| v \right \|_2}$ and $\theta = {\left \| v \right \|_2}$ are the axis and angle of rotation of $v$ as the 3-dimensional rotation vector representation of $R = R_s^TR_{GT}$, and knowing that the norm of the skew-symmetric matrix $[\hat{v}]_\times$ of unit vector is one, one can show that (\ref{eq:dis2}) is equal to $\theta$.

On the other hand, since the distance between $R_s$ and $R_{GT}$ can be represented as rotation matrix $R=R_s^TR_{GT}$ using (\ref{trace}), $\theta$ is equal to $cos^{-1}\left [ \frac{tr(R_s^TR_{GT})-1}{2} \right ]$. Therefore, the geodesic loss which is defined as the distance between two rotation matrices can be written as:
\begin{equation}
    Loss_{Geodesic} = d(R_s, R_{GT}) = cos^{-1}\left [ \frac{tr(R_s^TR_{GT})-1}{2} \right ]
    \label{eq:dis3}
\end{equation}
This is a natural \textcolor{black}{Riemannian} metric on the compact Lie group $SO(3)$. \textcolor{black}{Equations~(\ref{eq:logR}) and (\ref{eq:dis3}) are equivalent, so we calculate the geodesic loss using (\ref{eq:dis3}), as it is easier to implement. To use (\ref{eq:dis3}) we find the rotation matrices as described in Section~\ref{sec:Representation}.} In summary, training the networks involves iterations of back-propagation with the total loss function in (\ref{eq:total}) where translation loss is the MSE, and the rotation loss is calculated by (\ref{eq:MSE}) in the first stage and by (\ref{eq:dis3}) in the second stage. \textcolor{black}{This schedule is chosen because of the computational convexity advantage of MSE and the accuracy of the geodesic loss.} In our experiments each stage involved ten epochs. The details of the data and experiments are discussed next.

\section{Experiments}
\label{sec:results}
\subsection{Datasets}
\label{sec:datasets}
\textcolor{black}{The datasets used in this study contained 93 reconstructed T2-weighted MRI scans of fetuses, as well as T1- and T2-weighted MRI scans of 40 newborns. The newborn data was obtained from the first data release of the developing human connectome project~\cite{hughes2017developing}. The fetal MRI data was obtained from fetuses scanned at Boston Children's Hospital at a gestational age between 21 and 37 weeks (mean=30.1, stdev=4.6) on 3-Tesla Siemens Skyra scanners with 18-channel body matrix and spine coils. Written informed consent was obtained from all pregnant women research participants.} Repeated multi-planar T2-weighted single shot fast spin echo scans were acquired of the moving fetuses, ellipsoidal brain masks were automatically extracted based on the real-time algorithm in~\cite{salehi2018real}. The scans were then combined through slice-level motion correction and robust super-resolution volume reconstruction~\cite{gholipour2010robust,kainz2015fast}. Brain masks were generated on the reconstructed images using Auto-Net~\cite{salehi2017auto} and manually corrected in ITK-SNAP~\cite{yushkevich2006user} as needed.

Brain-extracted reconstructed images were then registered to a spatiotemporal fetal brain MRI atlas~\cite{gholipour2017normative} \textcolor{black}{at an isotropic resolution of $0.8mm^3$}. This registration was performed through the procedure described in~\cite{gholipour2017normative} and is briefly described here as it generated the set of fetal brain scans (all registered to the standard atlas space) used to generate ground truth data. First, a rigid transform $T_{w\rightarrow f}$ was found between the fetal head coordinates and the MRI scanner (world) coordinates by inverting the direction cosine matrix of one of the original fetal MRI scans that appeared in an orthogonal plane with respect to the fetal head (the idea behind this is that the MR technologist who prescribed scan planes identified and used the fetal head coordinates and did not use the world coordinates). Applying $T_{w\rightarrow f}$ to the image reconstructed in the world coordinates mapped it to the fetal coordinates; thus the oblique reconstructed image appeared orthogonal with respect to the fetal head after this mapping; which in-turn enabled a grid search on all orthogonal 3D rotations that could map this image to the corresponding age of the spatiotemporal atlas (fetal coordinates to atlas space). Multi-scale rigid registration was performed afterwards to fine tune the alignment.

\textcolor{black}{It should be noted that due to differences in the anatomy of different subjects at different ages and the templates, \textcolor{black}{the final alignments have an intrinsic level of uncertainty as an exact rigid alignment of two different anatomies is not well defined;} but since our goals are improved capture range and speed, in our analysis we are not sensitive to uncertainty in alignment of reference data. 
All images were manually controlled to ensure visually-correct alignment to the atlas space.}

\subsubsection{Training Dataset}
\label{sec:trainset}
\textcolor{black}{From the total database of 93 fetal MRI scans,} reconstructed T2-weighted images of 36 fetuses scanned at 28 to 37 weeks GA \textcolor{black}{were used all together to train one network.} Each image was 3D rotated and translated randomly and fed to the network. Since the rotation matrix was known the rotation vector was computed and used as the ground truth. Two different algorithms were used to randomly generate rotation matrices.

\textcolor{black}{For slice pose estimation training,} each input image randomly rotated around the $x, y,$ and $z$ axes between $\frac{-\pi}{2}$ and $\frac{+\pi}{2}$. This algorithm covered half of all possible orientations, and provided all different views in the training set. Therefore, for training the network, the separation of different views (i.e. axial, coronal, and saggital) was unnecessary. The reason that we did not span the whole space in this experiment is that 2D brain slices do not have enough information to separate between rotations that are $\pi$ radians away around arbitrary rotation vectors as predicting the 3D direction of the brain from a 2D slice is difficult due to the symmetrical shape of the brain. In order to choose input slices we randomly chose 30 slices from 66 percent of the middle slices, skipping the border slices that did not carry sufficient information for training.

\textcolor{black}{For volume pose estimation training} 
we used the algorithm proposed in~\cite[p.~355]{arvo2013graphics} to uniformly span the whole space. This algorithm mapped three random variables in the range $[0,1]$ onto the set of orthogonal $3\times3$ matrices with positive determinant; that is, the set of all 3D rotations. 
This algorithm generated uniformly distributed samples on unit sphere.

For the volume-to-volume training of the correction network \textcolor{black}{(referred to as the Correction-Net),} each moving image was randomly rotated around the $x, y,$ and $z$ axes between $\frac{-\pi}{6}$ to $\frac{+\pi}{6}$ and translated randomly in each direction between $-7$ to $7$ millimeters. The transformed image was then concatenated with its corresponding atlas image to form a 2-channel input to the network. The range of transform parameter variations was lower for this network as the objective of this network was to correct initial predictions made by other networks. \textcolor{black}{Initial translations between stack-of-slices or volumes and the template or reference were estimated using center-of-gravity matching and rotations were estimated by the pose estimation networks, therefore the Correction-Net was trained and tested on a limited range of transformation parameters; however, to evaluate the capacity of this network to learn to predict the entire parameter space for comparison purposes, we also trained it on the wider range of rotations similar to the 3D pose estimation network.  
We refer to this trained model as the 3DReg-Net in the results, as compared to the Correction-Net which was trained only to correct initial transformations.}

\textcolor{black}{Translation and rotation of the images were applied using one transformation and the resampling was done on-the-fly during training. Linear interpolation was used for resampling images for faster training. Scaling with random scale factors in the range of 0.95 and 1.05 were also used for data augmentation. The total number of generated training samples was $5,400,000$ slices for slice pose estimation and $180,000$ volumes for volume-to-volume registration. The number of epochs for each training step was set to $10$.}  

\subsubsection{Testing Datasets}
To test the performance and generalization properties of the trained models, three test sets were used: Test Set 1) reconstructed T2-weighted images of 40 fetuses with GA between 27 to 37 weeks that were not used in training\textcolor{black}{, as well as original T2-weighted slices of those scans}; Test Set 2) reconstructed T2-weighted images of 17 fetuses with GA between 22 and 26 weeks; as well as T2-weighted MRI scans of 7 newborns scanned at 38 to 44 weeks GA-equivalent age \textcolor{black}{(selected from the total number of 40 cases, to span the age range)}; and Test Set 3) T1-weighted MRI scans of those newborns. \textcolor{black}{There was no overlap between the test sets and the training set described in the previous subsection.}

\begin{figure}
    \centering
    \includegraphics[width=0.9\columnwidth]{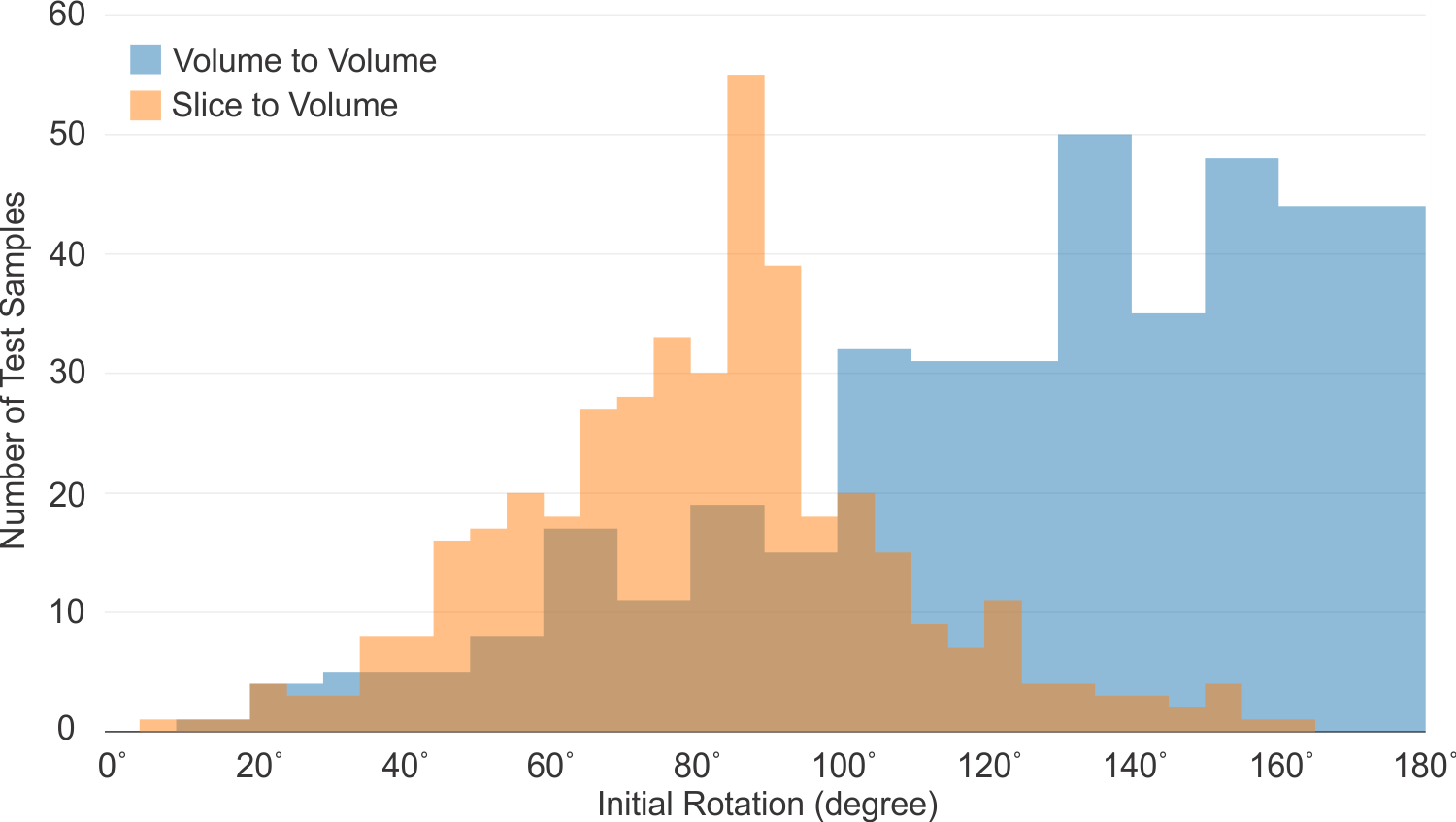}
    \caption{Histogram of distance from correct orientation in degrees in test sets. Orange shows slice samples created by using rotation around $x, y,$ and $z$ axes and blue shows volume samples created using the algorithm proposed in~\cite[p.~355]{arvo2013graphics}. This algorithm generated uniformly distributed samples on the sphere that resulted in these distributions of rotation angles in half-space \textcolor{black}{used in slice 3D pose estimation and full-space used in volume pose estimation}.}
    \label{fig:Histogram}
\end{figure}

On each 3D image 10 randomly generated rotation matrices were applied resulting in 400, 170, and 70 samples for each set. For each application, rotation matrices were generated through the same process used for the training data as discussed in section~\ref{sec:trainset}. Figure~\ref{fig:Histogram} shows the histogram of the synthetic rotations \textcolor{black}{for the slice and volume pose estimation experiments}. The $x$ axis shows the distance of the generated rotation matrix from the identity matrix in degrees.

\subsection{Intensity-Based Registration}
To compare the pose predictions made by our pose estimation CNNs, referred to as 3DPose-Net, with conventional intensity-based registration methods, we developed multiple variations of rigid registration for volume-to-volume registration (VVR) and slice-to-volume registration (SVR) \textcolor{black}{between images and age-matched templates}. For VVR comparisons, we developed the following programs: (i) VVR-GC: \textcolor{black}{A multi-scale approach was used for rigid registration with 3 levels of transform refinement. The transform was initialized using a Gravity Center (GC) matching strategy. A gradient-descent optimizer was used to maximize the normalized mutual information metric between the source and reference images}. (ii) VVR-PAA: the same as VVR-GC except that the transform was initialized using a moments matching and principal axis alignment approach. (iii) VVR-Deep: same as VVR-GC except that the transform was initialized using 3DPose-Net predicted transforms, without employing any other initialization strategy. For SVR comparisons, we developed two versions of the program: (i) SVR-GC: \textcolor{black}{A multi-scale approach was used for registration with 3 levels of transform refinement initialized with center-of-gravity matching, and gradient-descent maximization of normalized cross correlation.} (ii) SVR-Deep: same as SVR-GC except that the transform was initialized using 3DPose-Net predicted transforms. The learning rate for the optimization process was set lower in both VVR and SVR programs when they were initialized using 3DPose-Net predictions.

\subsection{Results}
We evaluated pose predictions obtained from the proposed methods in different scenarios. 
\subsubsection{Slice 3D Pose Estimation}
\begin{figure}
    \centering
    \includegraphics[width=\columnwidth]{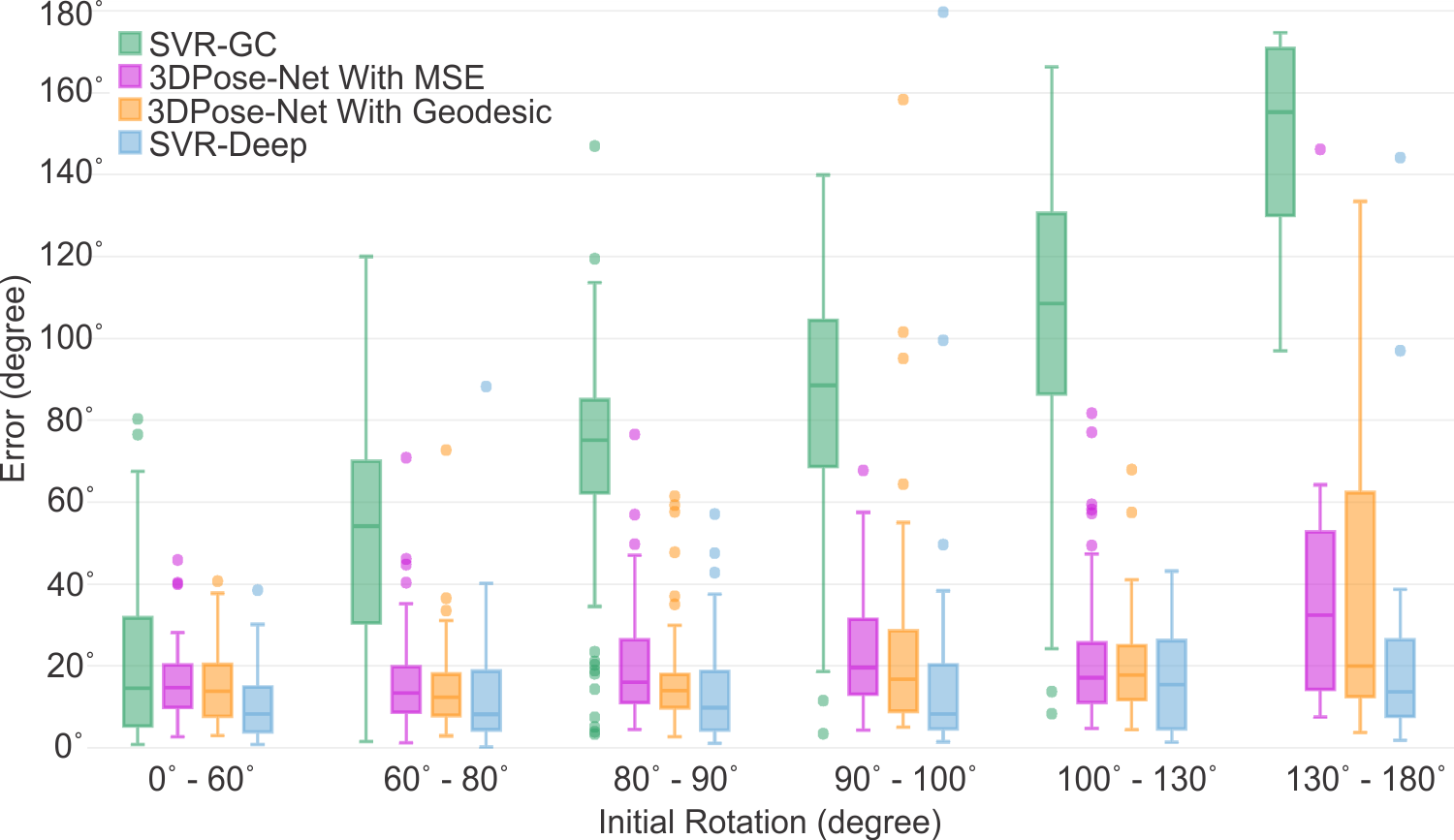}
    \caption{Errors (in degree) of \textcolor{black}{slice 3D pose estimation} tests. Median is displayed in boxplots; dots represent outliers outside 1.5 times the inter-quartile range of the upper and lower quartiles, respectively. Overall, these results show that the trained deep CNNs predicted the 3D pose of single slices very well despite the large range of rotations and significantly improved the performance of the optimization-based method (in SVR-deep). The performance of the 3DPose-Net fine tuned with geodesic loss was consistently superior to 3DPose-Net with MSE. Note that finding the 3D pose of subjects with different anatomy using a single slice is a difficult task.}
    \label{fig:S2V}
\end{figure}

As described in section~\ref{sec:method}, optimization-based SVR methods and the trained CNN were used for slice 3D pose estimation. \textcolor{black}{To investigate the influence of geodesic loss compared to MSE, after the model was trained with the MSE loss function, it was fine tuned for 10 more epochs, once with the MSE loss and once with the geodesic loss. In visualizing and comparing the results, test samples were distributed over 6 different bins according to their magnitude of rotation in a way that the number of samples in each bin was roughly equal. By this comparison we aimed to evaluate performance of methods in terms of their capture range.} It can be seen in Figure~\ref{fig:S2V} and Table~\ref{table:S2V} that 1) the geodesic loss improved the results. This improvement was significant in bins of $80-90^\circ$ and $90-100^\circ$ ; 2) the optimization based method without deep CNN initialization failed in most cases; and 3) the optimization-based method with deep initialization performed the best.

\begin{figure}
    \centering
    \includegraphics[width=\columnwidth]{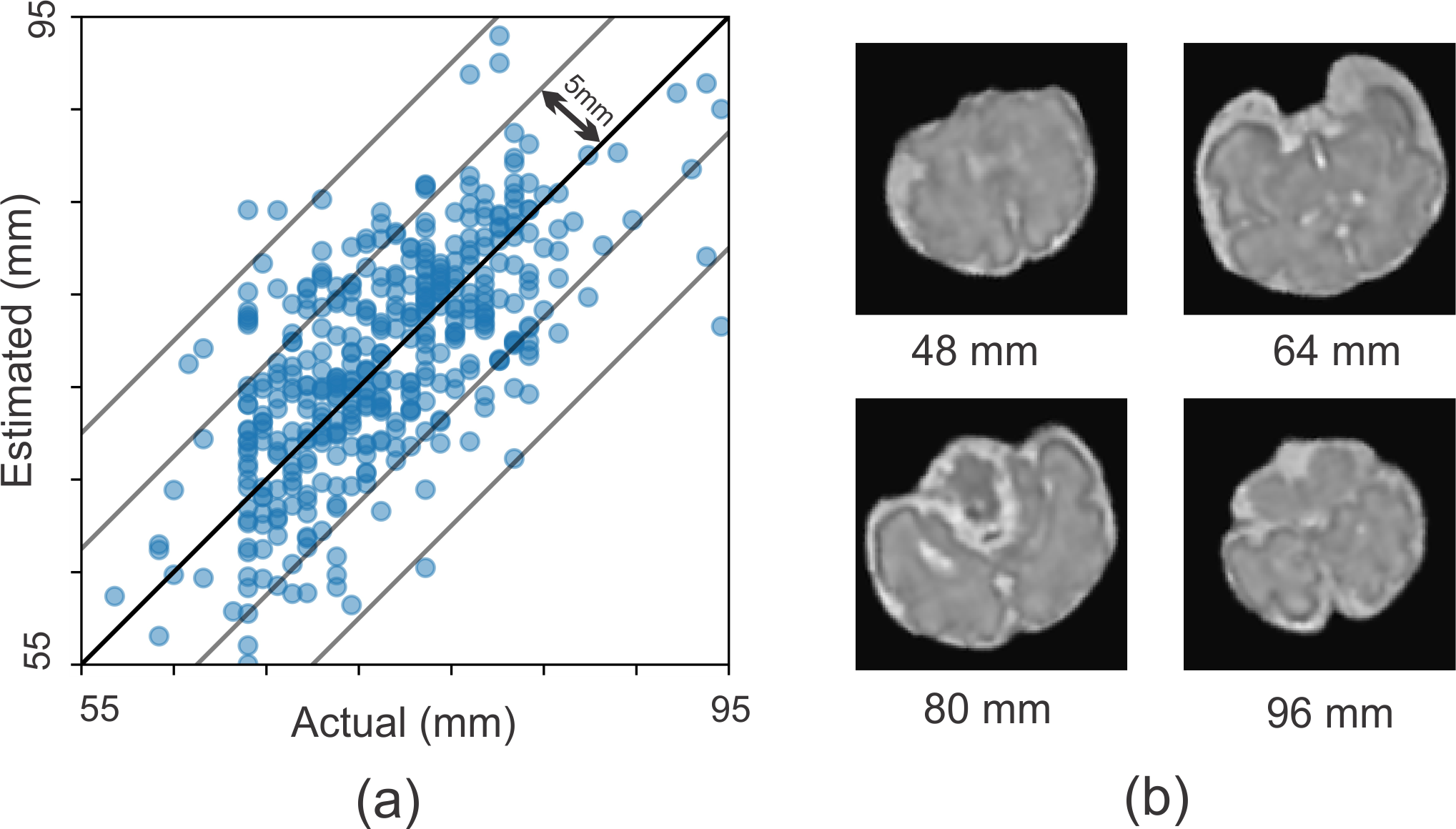}
    \caption{\textcolor{black}{Analysis of the slice location estimation: (a) estimated vs. actual slice locations in millimeters for the test data. Lines show $5mm$ and $10mm$ error margins. The error was $<5mm$ for the majority of slices, while it was higher for slices with limited features; (b) four sample slices of a test data.}}
    \label{fig:Slice}
\end{figure}

\textcolor{black}{Figure~\ref{fig:Slice}.a shows estimated physical location of slices compared to their actual location, with error lines (in mm). The estimation error of the majority of slices was below 5mm, while the error was higher for some slices especially for those closer to the boundary of the brain. Figure~\ref{fig:Slice}.b shows snapshots of four slices of one of the test cases. This figure shows the limited features available to the algorithm and the similarity of slices especially those closer to the boundaries. Learning to estimate slice locations for subjects with different anatomies scanned at different ages is challenging but can be augmented with slice-level motion tracking algorithms, such as~\cite{marami2017temporal}, when motion is fast and continuous.}  


\begin{table*}[h!]
\small
\centering
 \begin{tabular}{|c||c|c|c|c|c|c|} 
 \hline
  &  \multicolumn{6}{c|}{Slice 3D pose estimation error} \\
 \hline
Method & $0^{\circ} - 60^{\circ}$  & $60^{\circ} - 80^{\circ}$ & $80^{\circ} - 90^{\circ}$ & $90^{\circ} - 100^{\circ}$ & $100^{\circ} - 130^{\circ}$ & $130^{\circ} - 180^{\circ}$ \\
 \hline
SVR-GC & 21.36 ($\pm19.36$) & 50.62 ($\pm26.19$) & 70.16 ($\pm28.1$)& 83.28 ($\pm29.32$) & 105.64 ($\pm34.68$) & 147.69 ($\pm24.80$) \\ 
 \hline
 \color{black}
3DPose-Net (MSE) & 15.73 ($\pm8.3$) & 15.9 ($\pm10.6$) & 20.9 ($\pm13.23$)& 22.74 ($\pm12.87$) & 21.66 ($\pm16.13$) & 38.45 ($\pm32.10$) \color{black} \\ 
 
 \hline
3DPose-Net (Geodesic) & 15.1 ($\pm8.25$) & 14.05 ($\pm9.16$) & 16.18 ($\pm11.44$)& 24.33 ($\pm26.5$) & 19.8 ($\pm11.44$) & 36.47 ($\pm34.25$) \\ 
 \hline
SVR-Deep & \textbf{10.23} ($\pm7.77$) & \textbf{12.32} ($\pm11.35$) & \textbf{13.08} ($\pm11.8$)& \textbf{17.6} ($\pm26.9$) & \textbf{16.19} ($\pm11.45$) & \textbf{26.85} ($\pm35.5$) \\ 
 \hline
\end{tabular}
\caption{Mean and standard deviation of errors in degree for different algorithms on 400 samples generated from 40 different subjects. The results show that using optimization based algorithms with deep CNN predicted priors significantly reduced the errors. Note the magnitude of synthetic rotations in the first row. Finding the pose of subjects with different anatomy using a single slice is a difficult task.}
\label{table:S2V}
\end{table*}

\begin{figure*}
    \centering
    \includegraphics[width=\textwidth]{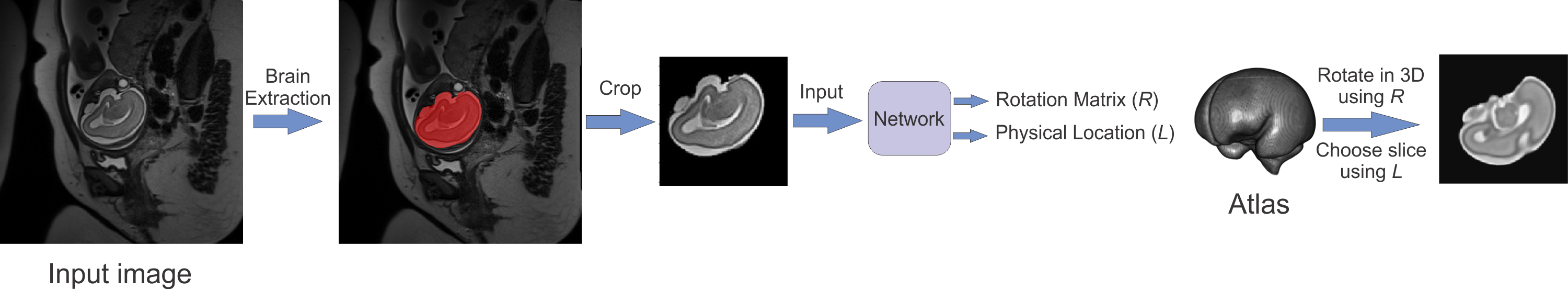}
    \caption{\textcolor{black}{The pipeline used to test the trained 3DPose-Net on original fetal MRI slices in the test set. The fetal brain was extracted using a previously published technique~\cite{salehi2018real}, and fed into 3DPose-Net. The transformation composed of the estimated slice pose and location was applied to the age-matched atlas to find the corresponding slice from the atlas.}}
    \label{fig:Actual2D}
\end{figure*}

\begin{figure}
    \centering
    \includegraphics[width=0.95\columnwidth]{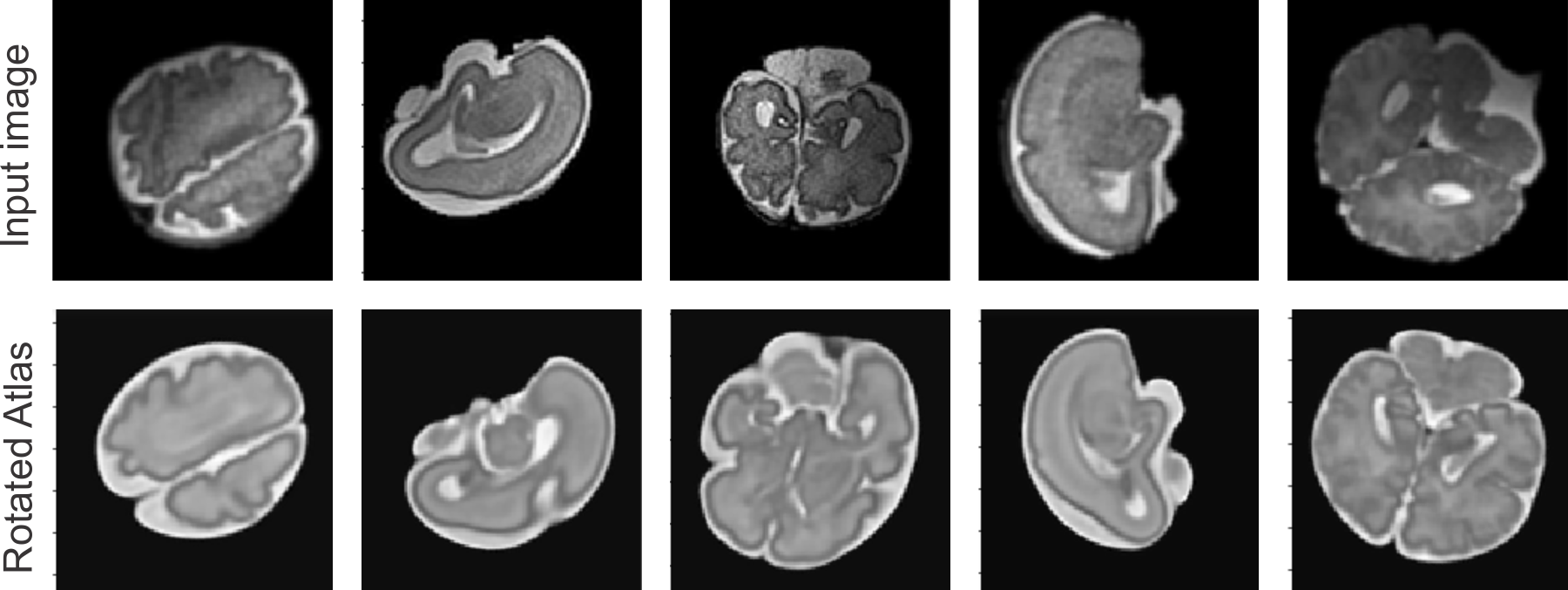}
    \caption{\textcolor{black}{Five sample slices from five fetuses in different ages. The first row shows cropped version of brain-extracted slices using the pipeline shown in figure~\ref{fig:Actual2D}, and the second row shows the corresponding rotated atlas slices obtained from 3DPose-Net slice pose estimations.}}
    \label{fig:Actual2D_cmp}
\end{figure}

\textcolor{black}{We also tested the trained model on original slices from T2-weighted stack-of-slices of the test subjects. the goal in this experiment, which is shown in Figure~\ref{fig:Actual2D}, was to find the corresponding 3D pose and location of the fetal brain in an input slice in the atlas space. The fetal brain was extracted in each slice using~\cite{salehi2018real} and was provided as input to the trained slice 3DPose-Net. A transformation composed from the estimated pose and location of the slice was applied to the corresponding age of the spatiotemporal fetal brain MRI atlas~\cite{gholipour2017normative} to obtain the corresponding atlas slice. Representative results of original slices and the corresponding estimated slices from the atlas are shown for five samples from different fetuses in Figure~\ref{fig:Actual2D_cmp}.} 


\subsubsection{\textcolor{black}{Volume 3D Pose Estimation}}
\begin{figure}
    \centering
    \includegraphics[width=\columnwidth]{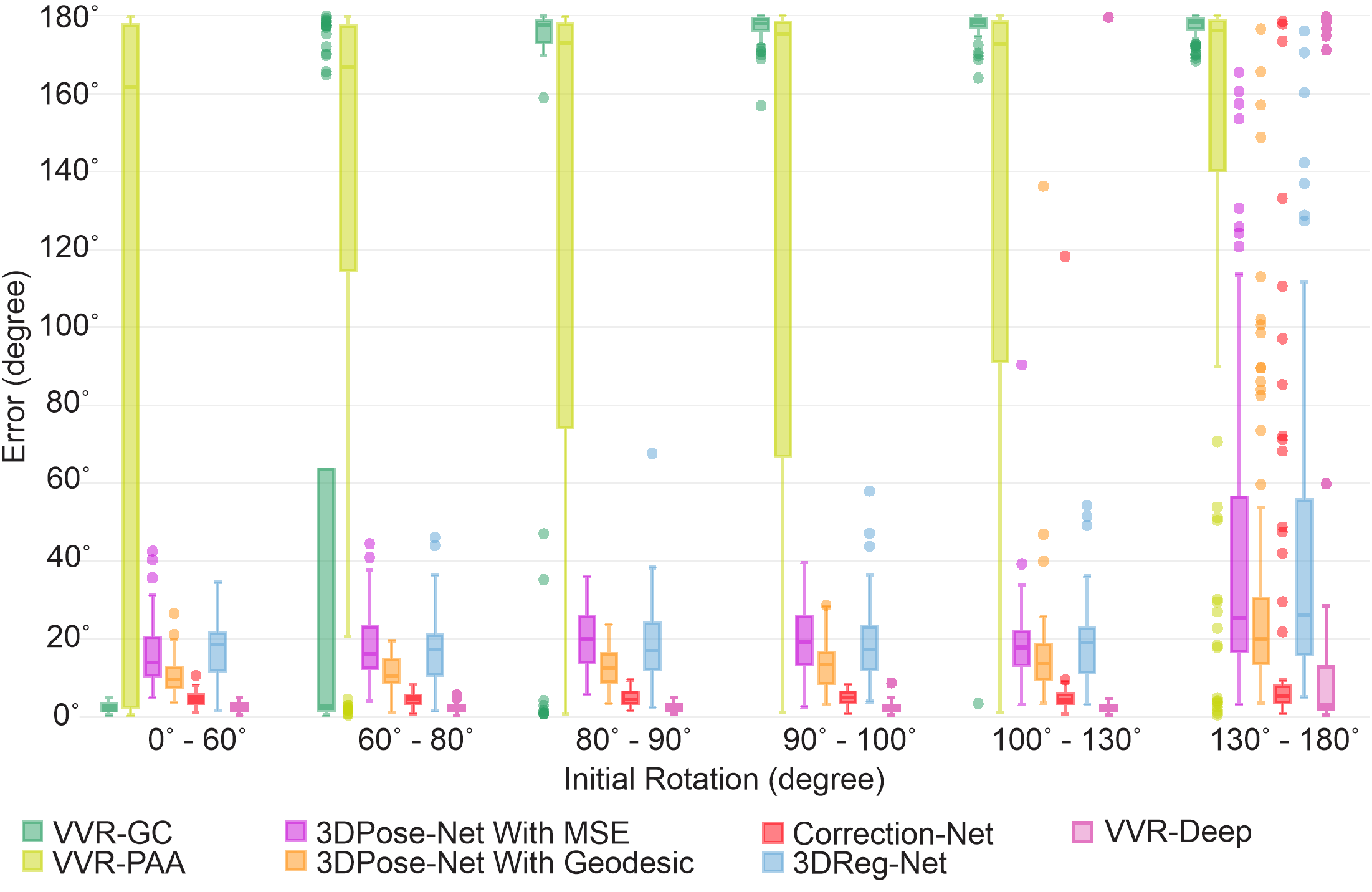}
    \caption{Errors (in degree) of the volume-to-volume registration experiments. Median is displayed in boxplots; dots represent outliers outside 1.5 times the interquartile range of the upper and lower quartiles, respectively. Overall, these results show that the VVR-Deep (VVR initialized with deep predictions) generated the most accurate results and the least number of failures. The Correction-Net performed comparably in most regions while being significantly faster than VVR-GC. VVR-GC performed very well for small rotations ($0-60^\circ$) but failed for almost all rotations $>80^\circ$. VVR-PAA did not show a robust performance either and failed in many cases. 3DPose-Net with geodesic loss showed significant improvement over the 3DPose-Net with the MSE loss.}
    \label{fig:V2V}
\end{figure}

\textcolor{black}{In the volume-to-volume rigid registration scenario, 6 different algorithms were compared: VVR-GC, VVR-PAA, 3DPose-Net with MSE, 3DPose-Net with geodesic loss, Correction-Net, 3DReg-Net, and VVR-Deep.} Figure~\ref{fig:V2V} shows that 1) the VVR-GC performed very well for rotations between $0-60^\circ$ but it failed for almost all samples with rotations $>80^\circ$ as it converged to the wrong local minima; 2) by using the principal axis initialization, the VVR-PAA significantly improved the performance for $>80^\circ$ but again failed for the majority of samples with rotations, and it resulted in a huge loss in performance (compared to VVR-GC) in $0-60^\circ$ as it incorrectly shifted the initial point to the region of a wrong local minimum. 3) The trained deep CNN models all performed well as they showed much lower number of failures. The geodesic loss showed significant improvement over the MSE loss; and the Correction-Net performed the best with only a very small fraction of failures in the $130-180^\circ$ range of rotations. 4) VVR-Deep, which is the optimization-based registration initialized by deep pose estimation generated the most accurate results and the minimum number of failures. Table~\ref{table:V2V} shows that VVR-Deep performed the best, while Correction-Net results were also comparable, especially as Correction-Net based registration is real-time and several orders of magnitude faster than the VVR-Deep registration. The average runtime of methods is discussed in Section~\ref{sec:testtimes}. 

\begin{table*}[h!]
\small
\centering
 \begin{tabular}{|c||c|c|c|c|c|c|} 
 \hline
  &  \multicolumn{6}{c|}{Volume-to-Volume} \\
 \hline
Method & $0^{\circ} - 80^{\circ}$  & $80^{\circ} - 110^{\circ}$ & $110^{\circ} - 130^{\circ}$ & $130^{\circ} - 145^{\circ}$ & $145^{\circ} - 160^{\circ}$ & $160^{\circ} - 180^{\circ}$ \\
 \hline
VVR-GC & \textbf{2.42} ($\pm1.28$) & 45.39 ($\pm73.74$) & 149.91 ($61.97$)& 177.0 ($3.62$) & 174.87 ($\pm21.4$) & 177.2 ($\pm2.73$) \\ 
\hline
VVR-PAA & 95.54 ($\pm84.91$) & 131.1 ($\pm66.99$) & 128.44 ($71.16$)& 129.68 ($71.36$) & 131.15 ($\pm67.34$) & 141.44 ($\pm62.88$) \\ 
 \hline
 \color{black}
3DPose-Net (MSE) & 16.28 ($\pm7.99$) & 17.85 ($\pm8.46$) & 20.06 ($\pm7.59$)& 19.5 ($\pm8.88$) & 18.93 ($\pm11.49$) & 45.38 ($\pm41.32$) \color{black} \\ 
 \hline
 
3DPose-Net (Geodesic) & 10.08 ($\pm4.42$) & 11.44 ($\pm3.97$) & 12.43 ($\pm4.56$)& 13.46 ($\pm5.98$) & 16.46 ($\pm16.57$) & 34.19 ($\pm37.54$) \\ 
 \hline
Correction-Net & 4.54 ($\pm1.66$) & 4.45 ($\pm1.73$) & 4.83 ($\pm1.85$)& 4.82 ($\pm1.79$) & 6.33 ($\pm13.99$) & \textbf{19.42} ($\pm38.66$) \\ 
 \hline
  \color{black}
3DReg-Net & 18.21 ($\pm7.77$) & 17.53 ($\pm8.67$) & 18.80 ($\pm10.17$)& 18.65 ($\pm10.29$) & 19.57 ($\pm10.39$) & 43.88 ($\pm39.94$) \color{black} \\ 
 \hline
  
VVR-Deep & \textbf{2.42} ($\pm1.28$) & \textbf{2.35} ($\pm1.08$) & \textbf{2.43} ($\pm1.19$)& \textbf{2.36} ($\pm1.33$) & \textbf{4.84} ($\pm21.68$) & 20.44 ($\pm46.81$) \\ 
 \hline
\end{tabular}
\caption{Mean and standard deviation of the errors in degree for different algorithms on 400 samples generated from 40 fetuses from the test set. The results show that VVR-Deep (optimization-based registration initialized with 3DPose-Net predictions) performed best. The correction network results were comparable.}
\label{table:V2V}
\end{table*}

\begin{figure}
    \centering
    \includegraphics[width=0.95\columnwidth]{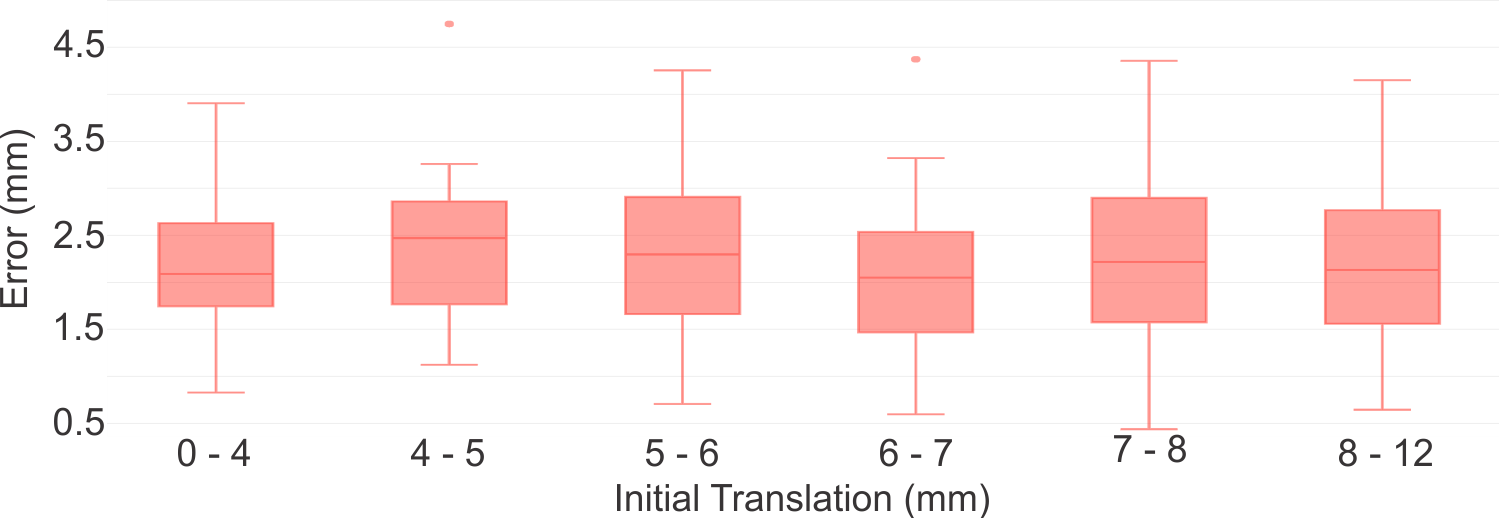}
    \caption{Errors (in mm) in volume-to-volume registration using the Correction-Net. Median is displayed in boxplots; dots represent outliers outside 1.5 times the interquartile range of the upper and lower quartiles, respectively. Note that there is an intrinsic level of inaccuracy in rigid registration of different brains as the exact alignment of two anatomies with different shapes and sizes cannot be achieved with rigid registration.}
    \label{fig:translation}
\end{figure}

Figure~\ref{fig:translation} shows the translation error of the correction network. The error is calculated as the distance of true translation vector and the predicted one. The initial translation was calculated as the distance of the input image to the atlas location. Note that all errors reported here including the translation errors are between images of different subjects and atlases, so there is an intrinsic level of uncertainty in alignment as the exact alignment of two different anatomies (with different size and shape) using rigid registration is not well defined.

\begin{figure}
    \centering
    \includegraphics[width=\columnwidth]{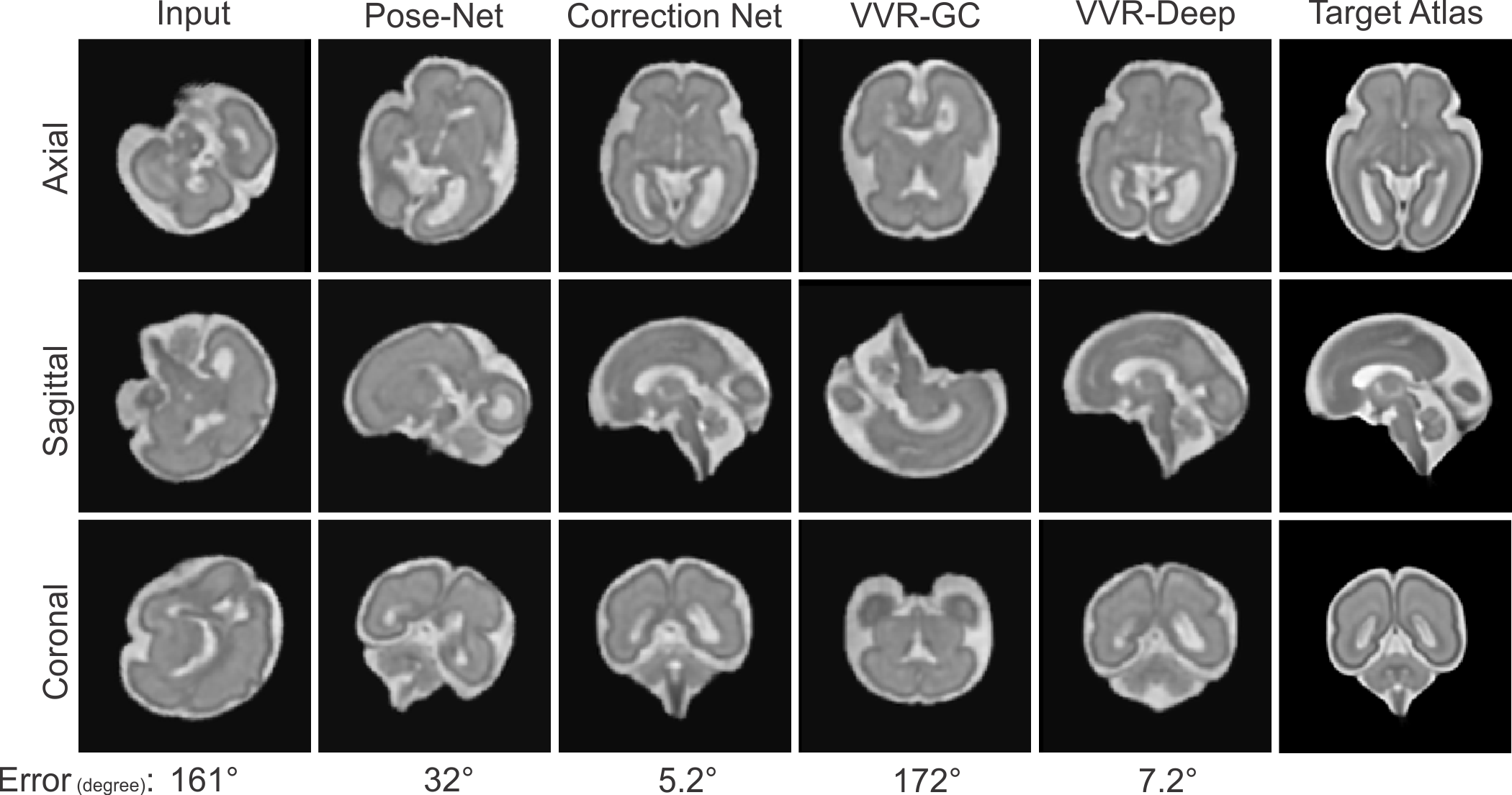}
    \caption{Three views of resampled images using transformations estimated with different algorithms for volume-to-volume registration. The first column is the synthetically-rotated input and the last column is the target atlas image. The geodesic distance errors of the algorithms are shown in degrees underneath each column. The correction network worked best in this example. The optimization-based registration, VVR-GC, failed as it converged to a local minimum, whereas it worked well when initialized with the 3DPose-Net predicted transformation (VVR-Deep).}
    \label{fig:3DVis}
\end{figure}

Figure~\ref{fig:3DVis} shows the results of different algorithms on an example from the volume-to-volume registration tests. All algorithms tried to register the brain of this fetus (with mild unilateral ventriculomegaly) to the corresponding age of the atlas on the right. The first column is the input with synthetic rotation. As the rotation was more than $90^\circ$, VVR-GC failed due to the non-convex similarity loss function. Without deep initialization this algorithm converged to the wrong local optimum which resulted in a flipped version of the correct orientation (the forth column). The second and third columns show the results of the 3DPose-Net and the Correction-Net. The geodesic distance errors (in degrees) of each algorithm are given underneath each column. For this example, the correction network generated the most accurate results. 

\subsubsection{Generalization property of the trained models}
\begin{figure*}
    \centering
    \includegraphics[width=\textwidth]{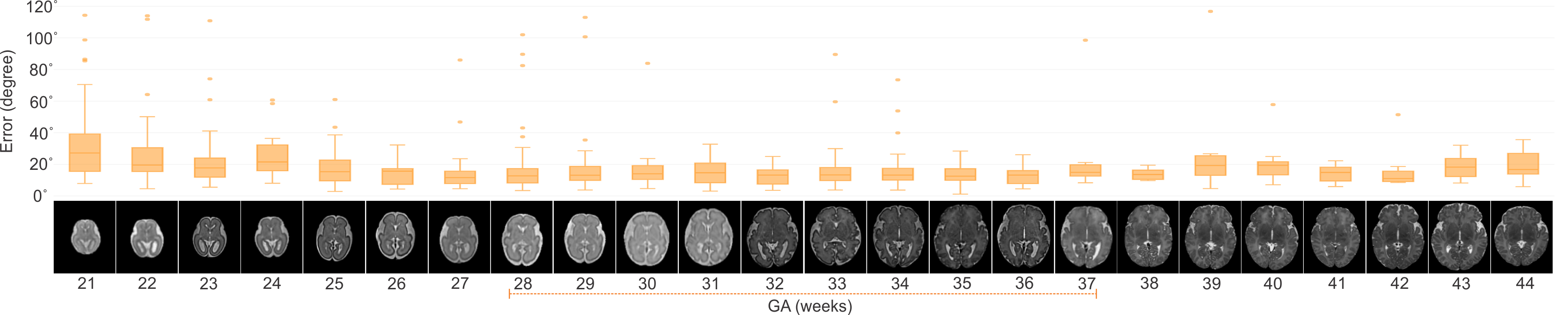}
    \caption{Generalization property of the pose network (3DPose-Net) on different ages and different scan settings (fetal vs. newborn scans). The network was trained using fetal samples scanned between 27-37 weeks (orange dashed line), and tested on fetal samples scanned at 22-26 weeks and newborns scanned \textit{ex-utero} at 38-44 weeks (Test Set 2). The variation in size and shape of the brain is high as the brain develops very rapidly throughout 21-30 weeks GA. This figure shows that while there were large systematic differences between the train and test data, the network generalized well to estimate the pose.}
    \label{fig:GA}
\end{figure*}

An important question that is frequently asked about learning-based methods such as the ones developed in this study concerns their generalization performance: can they generalize well for new test data, possibly with different features? In this section, we aimed to investigate the generalization property of our trained models. For this, we carried out two sets of experiments, with Test Sets 2 and 3:

First, we added Test Set 1 to Test Set 2 to investigate the generalization of the algorithm for fetal brains at ages other than those used in the training set (younger fetuses at 22-27 weeks GA and newborn brains scanned at 38-44 weeks GA-equivalent age scanned in different, \textit{ex-utero} scan settings). We recall that the training dataset only contained fetuses scanned at 28-37 weeks. The brain develops very rapidly especially throughout the second trimester of pregnancy, therefore the difference in brain size and shape between these test sets and the training set was significant. The images underneath the box plots in Figure~\ref{fig:GA} show sample slices for different ages. By simply using a scale parameter that was calculated by the size ratio of atlases at different ages, we scaled the images and fed them into the network. Box plots of the estimated pose error in different ages in Figure~\ref{fig:GA} showed that the network generalized very well over all age ranges and for different scan settings. It is, however, seen that the average and median errors slightly increased towards the lower age range as the anatomy became significantly different from the anatomy of the training set.  

\begin{figure}
    \centering
    \includegraphics[width=0.9\columnwidth]{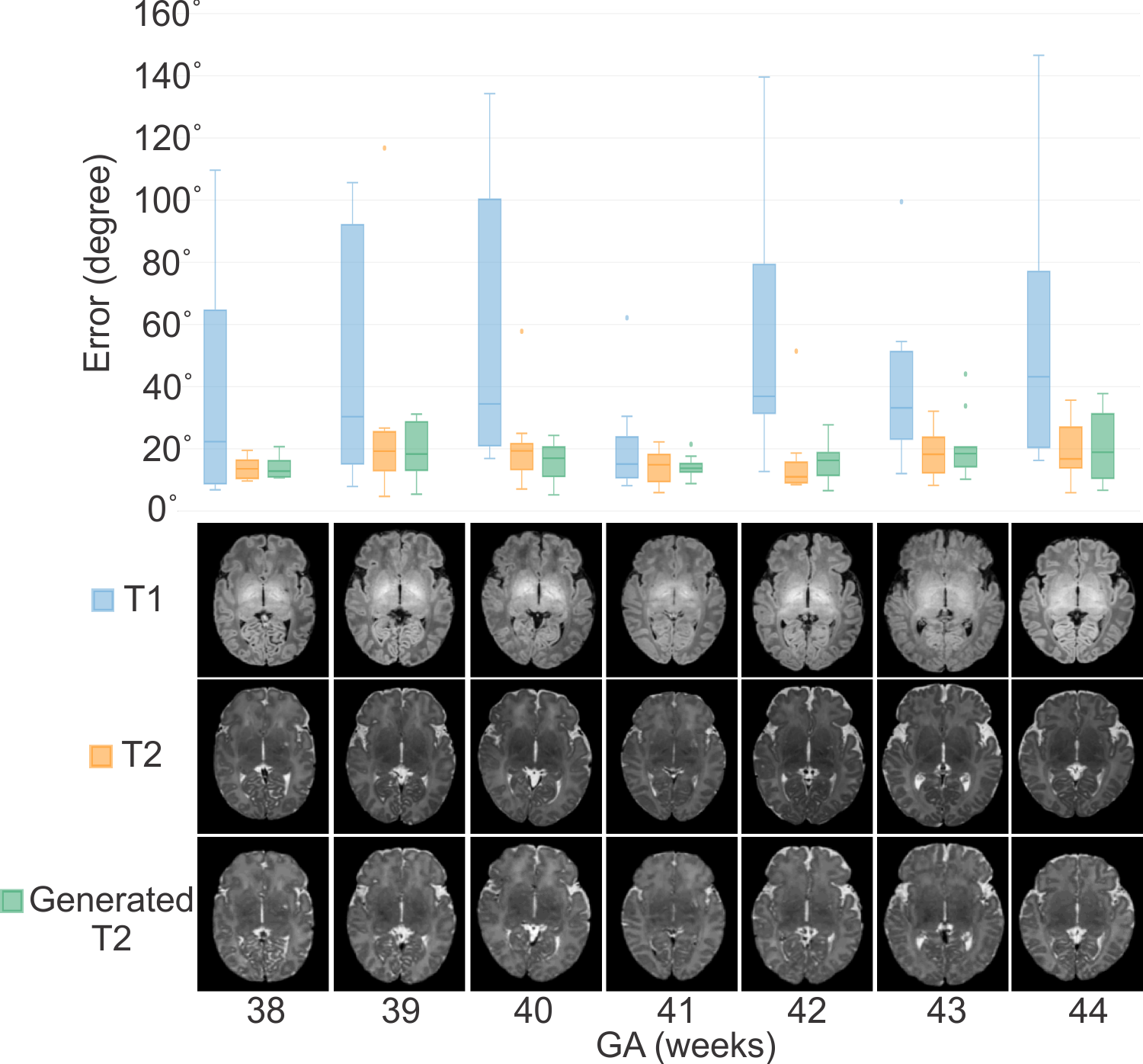}
    \caption{Generalization of the pose estimation network over different modalities. The network is trained on reconstructed T2-weighted images and is not generalized well on T1 modalities (blue boxes). However, the network estimated the pose very well using generated T2-weighted images (last row).}
    \label{fig:T1toT2}
\end{figure}

In our second experiment on generalization, we investigated generalizability of the networks over different modalities. To investigate whether the 3DPose-Net could generalize on T1-weighted newborn MRI scans while trained only on reconstructed T2-weighted scans of fetuses, we applied our volume-to-volume registration test pipeline to T1-weighted scans of 7 newborns (70 samples in total) in Test Set 3. Figure~\ref{fig:T1toT2} shows the results of applying the trained model on T1-weighted scans (blue box plots) compared to T2-weighted scans (orange box plots) with exact same random rotations. While 3DPose-Net still performed better than VVR-GC and VVR-PAA (compared to Figure~\ref{fig:V2V}), it did not generalize well on T1-weighted scans.

\textcolor{black}{To solve this issue through pre-processing, we developed an image contrast transfer algorithm based on a conditional generative adversarial network (GAN)~\cite{isola2017image}. Details of this algorithm can be found in Appendix~\ref{appendix}. By training a conditional GAN in this approach, we transferred T1-weighted images to T2-weighted images.} The results of the \textcolor{black}{transferred} T2-weighted images are shown in the last row of Figure~\ref{fig:T1toT2}. The pose error box plots in this figure show that the image contrast transfer from T1 to T2-weighted images and using the generated T2 images as input to the pose network significantly decreased the pose estimation error. In fact the trained \textcolor{black}{T2-weighted image} generator can be used as an input cascade to the 3DPose-Net or Correction-Net so that they can be directly used to register T1-weighted newborn brain images without being trained in this domain. Note that no reference data (aligned to an atlas) was needed for T1-weighted scans to train the conditional GAN except a set of paired T1 and T2 scans in the subject space that was easy to obtain. A similar approach can be taken to further expand the generalization domain of the trained pose estimation networks, for example to adult brains. In this work we had access to paired T1 and T2 images. In case in any other application paired images are not accessible between two domains, cycleGAN~\cite{zhu2017unpaired} can be used.

\subsubsection{Testing times}
\label{sec:testtimes}
\textcolor{black}{Table~\ref{table:Time} shows the average testing time (in milliseconds) for the algorithms developed in this study, measured on a GPU, which shows that all the deep learning based algorithms were real time. The test time difference between 3DPose-Net and the Correction-Net was because of a resampling operation on the image between the two stages of the Correction-Net, which took about 80 milliseconds. For comparison, we note that efficient implementations of intensity-based optimization methods for VVR typically require about 5ms per iteration of optimization (for 1M voxel samples) on GPUs, and about 10ms per iteration of optimization through symmetric multiprocessing~\cite{shams2010survey}. Depending on the range of rotations and translations, which may require a multi-scale registration approach and between 10-100 iterations of optimization, these algorithms may take between 50ms and several seconds if implemented efficiently on appropriate hardware. All pre-processing steps prior to the application of 3DPose-Net and Correction-Net were also real-time. \textcolor{black}{All experiments were done on an NVIDIA Geforce GTX 1080 GPU}. This includes the center-of-gravity matching to estimate initial translations, as well as scaling and the application of the conditional GAN. The average test time for the conditional GAN was $\sim50$ms. This analysis shows that the techniques proposed in this paper can improve the performance and capture range of massively parallel implementations of optimization-based registration algorithms~\cite{shams2010survey} for real-time applications such as image-based surgical navigation and motion-robust imaging.}

\begin{table}[h!]
\small
\centering
 \begin{tabular}{|c||c|c|} 
 \hline
Method & Volume & Slice \\
 \hline
3DPose-Net & $<5$ ms & $<5$ ms \\
 \hline
Correction-Net & $<100$ ms & - \\ 
 \hline
Conditional GAN & $\sim50$ ms & - \\
 \hline
\end{tabular}
\caption{\textcolor{black}{Average testing times (in milliseconds) for the methods in this study tested on an NVIDIA GeForce GTX 1080 (Pascal architecture). Given typical MRI slice acquisition times that vary between 50 to 2000 ms, these computation times enable real-time 3D pose estimation and registration.}}
\label{table:Time}
\end{table}

\section{Discussion}
\label{sec:discussion}
In this work we trained deep CNN regression models for 3D pose estimation of anatomy based on medical images. Our results show that deep learning based algorithms not only can provide a good initialization for optimization-based methods to improve the capture range of slice-to-volume registration, but also can be directly used for robust volume-to-volume rigid registration in real time. Using these learning-based methods along with accelerated optimization-based registration methods will provide powerful registration systems that can capture almost all possible rotations in 3D space.

Our networks composed of feature extraction layers and regression heads at the output layer. Using $\pi tanh$ non-linearity at the regression layer mimicked the behaviour of the angle-axis representation of the rotation matrix, where the geodesic loss was used as a bi-invariant, natural \textcolor{black}{Riemannian} distance metric for the space of 3D rotations. Compared to MSE on rotation vectors, our results showed that the geodesic loss led to significantly improved performance especially in 3D when images contained sufficient information for pose estimation.

By using a two step approach, where the 3D pose of an object (anatomy) is first approximately found in a standard (atlas) space, and then fed along with a reference image as two channels of input to a regression CNN (the correction network), accurate inter-subject rigid registration can be achieved in real-time for all ranges of rotation and translation. Initial translations may be achieved also in real-time through center of gravity matching.

One of the main concerns with learning based methods is their generalization property when they face test images with features that are different from the training set. This would be more important in medical imaging studies as the number of training samples is rather limited. In this study, to evaluate the generalization of the trained models over different ages, as the shape and size of the brain aggressively changes in early gestational weeks, we intentionally trained the network on older fetuses and tested it on younger ones. We only used a pre-defined scale parameter inferred from the gestational age based on a fetal brain MRI atlas.

We also tested the trained models on brain MRI scans of newborns which were obtained in a completely different setting, with head coils for \textit{ex-utero} imaging. While the trained models worked very well for T2-weighted brain scans of newborns at 38-44 weeks, we challenged the trained models by testing T1-weighted MRI scans of newborn brains. For the T1-weighted scans the performance of the networks dropped significantly; but we showed that by using a GAN based technique that learned to translate T1-weighted images into T2-like images, and feeding the outputs into the trained regression CNNs, we achieved great performance for T1-weighted images as well. To achieve this, we designed and trained an image-to-image translation GAN from pairs of T1 and T2 images of newborn subjects in a training set; and used it as a real-time pre-processing step for T1-weighted scans before they were fed into the pose estimation networks. In fact, with the conditional GAN algorithm, many of the learning based algorithms can be generalized over different modalities as long as some paired images are provided for training.

\section{Conclusion}
\textcolor{black}{We developed and evaluated deep pose estimation networks for slice-to-volume and volume-to-volume registration. In learning-based image-to-template (standard space) registration scenarios,} the proposed methods provided very fast (real-time) registration with a wide capture range on the space of all plausible 3D rotations, and provided good initialization for current optimization based registration methods. While the current highly-evolved multi-scale optimization-based methods that use cost functions such as mutual information or local cross correlation can converge to wrong local minima due to non-convex cost functions, our proposed CNN-based methods learn to predict 3D pose of images based on their features. A combination of these techniques and accelerated optimization-based methods can dramatically enhance the performance of imaging devices and image processing methods of the future.

\appendices
\counterwithin{figure}{section}
\section{}
\label{appendix}

\textcolor{black}{In the image contrast transfer algorithm we trained a conditional generative adversarial network (cGAN)~\cite{isola2017image} to simultaneously learn the mapping from T1- to T2-weighted images and a loss function to learn this mapping. Figure~\ref{fig:T1toT2Net} shows the pipeline to train the adversarial network on T1 and T2 image pairs. In this algorithm two networks, a generator ($G$) and a discriminator ($D$), were trained simultaneously in a way that $G$ tried to generate T2-like images from the T1-weighted scans, and $D$ tried to distinguish real from fake (synthetically-generated) T2-weighted image contrast in \{T1, T2\} pairs. To train these networks the following objective was used, where $z$ was random noise vector:
\begin{equation}
    G^* = arg\amin_G\amax_D \ L_{cGAN}(G,D) + \lambda L_{L1}(G)
\end{equation}
where the loss function of the cGAN, $L_{cGAN}(G,D)$, was defined as:
\begin{equation}
\begin{split}
    L_{cGAN}(G,D) = \  & \mathbb{E}_{T1, T2}[log\ D(T1,T2)]+ \\ 
    & \mathbb{E}_{T1, z}[log(1-D(T1,G(T1,z)))]
\end{split}
\end{equation}
and the distance between the generated and real T2 scans in the training set were calculated by the $L1$-norm to encourage generating sharper images:
\begin{equation}
    L_{L1}(G) =  \mathbb{E}_{T1, T2, z}[||T2-G(T1,z)||_1]
\end{equation}
\begin{figure}
    \centering
    \includegraphics[width=\columnwidth]{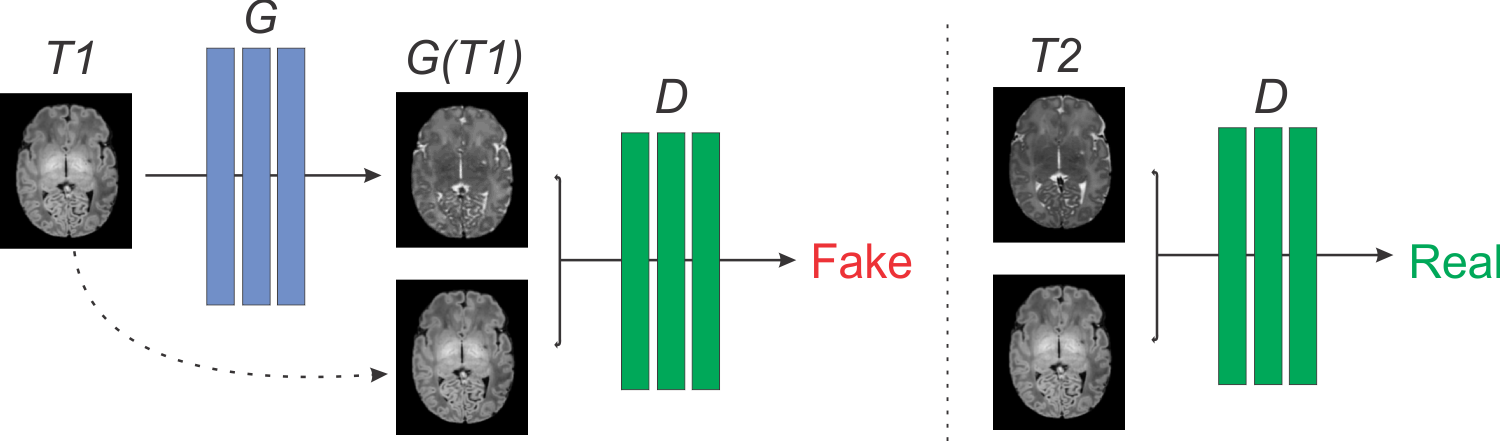}
    \caption{Training a conditional GAN to map T1$\rightarrow$T2 weighted images based on the approach in~\cite{isola2017image}. The discriminator, $D$, learns to classify fake (synthesized by the generator) and real \{T1, T2\} pairs. The generator, $G$, learns to fool the discriminator. Unlike an unconditional GAN, both the generator and discriminator observe the input T1-weighted image.}
    \label{fig:T1toT2Net}
\end{figure}
To train the conditional GAN networks we used 33 pairs of T1 and T2-weighted newborn brain images (of the subjects not used in the test set) resulting in 3300 paired slices. These images were used for training only. We then tested the trained $G$ on the test set of 7 newborn brain images.
}


\ifCLASSOPTIONcaptionsoff
  \newpage
\fi

\bibliographystyle{IEEEtran}

\bibliography{Ref}
\end{document}